\pdfoutput=1

\documentclass[preprint,review, 12pt]{elsarticle}

\usepackage{amssymb}

\usepackage{graphicx}
\usepackage{comment}
\usepackage{amsmath,amssymb} 
\usepackage{color}

\usepackage[american]{babel}

\usepackage{multirow}
\newcommand{\tabincell}[2]{
\begin{tabular}{@{}#1@{}}#2\end{tabular}
}
\usepackage{makecell}
\usepackage{xcolor}
\usepackage{colortbl}
\usepackage{url}
\usepackage{listings}
\definecolor{codegreen}{rgb}{0,0.5,0}
\definecolor{codeblue}{rgb}{0.25,0.5,0.5}
\definecolor{codegray}{rgb}{0.6,0.6,0.6}
\usepackage{color}

\definecolor{dkgreen}{rgb}{0,0.6,0}
\definecolor{gray}{rgb}{0.5,0.5,0.5}
\definecolor{mauve}{rgb}{0.58,0,0.82}

\usepackage{bm}

\usepackage[ruled,linesnumbered]{algorithm2e}

\journal{Pattern Recognition}

\begin{document}

\begin{frontmatter}



\title{MASTER: Multi-Aspect Non-local Network for \\Scene Text Recognition\tnoteref{copyright}}


\author[PINGAN]{Ning Lu\fnref{fn1}}
\ead{jiangxiluning@gmail.com}

\author[PINGAN,XZMU]{Wenwen Yu\corref{corresponding_author}\fnref{fn1}}
\ead{yuwenwen62@gmail.com}

\author[PINGAN]{Xianbiao Qi}
\ead{qixianbiao@gmail.com}
\author[PINGAN]{Yihao Chen}
\ead{o0o@o0oo0o.cc}
\author[XZMU]{Ping Gong}
\ead{gongping@xzhmu.edu.cn}
\author[PINGAN]{Rong Xiao}
\ead{xiaorong283@pingan.com.cn}
\author[HUST]{Xiang Bai}
\ead{xbai@hust.edu.cn}

\cortext[corresponding_author]{\ Corresponding author.}
\fntext[fn1]{\ Co-first authors.}

\tnotetext[copyright]{\ https://doi.org/10.1016/j.patcog.2021.107980 }
\tnotetext[copyright1]{\ \copyright\ 2021 The Author(s). This manuscript version is made available under the CC-BY-NC-ND 4.0 license http://creativecommons.org/licenses/by-nc-nd/4.0/}

\address[PINGAN]{Visual Computing Group, Ping An Property and Casualty Insurance Company, Shenzhen, China}
\address[XZMU]{School of Medical Imaging, Xuzhou Medical University, Xuzhou, China}
\address[HUST]{School of Electronic Information and Communications, Huazhong University of Science and Technology, Wuhan, China}

\begin{abstract}
  Attention-based scene text recognizers have gained huge success, which leverages a more compact intermediate representation to learn 1d- or 2d- attention by a RNN-based encoder-decoder architecture. However, such methods suffer from \textbf{attention-drift} problem because high similarity among encoded features leads to attention confusion under the RNN-based local attention mechanism. Moreover, RNN-based methods have low efficiency due to poor parallelization. To overcome these problems, we propose the MASTER, a self-attention based scene text recognizer that 
  (1) not only encodes the input-output attention but also learns self-attention which encodes feature-feature and target-target relationships inside the encoder and decoder and
  (2) learns a more powerful and robust   intermediate representation to spatial distortion, and
   (3) owns a great training efficiency because of high training parallelization and a high-speed inference because of an efficient memory-cache mechanism.
Extensive experiments on various benchmarks demonstrate the superior performance of our MASTER on both regular and irregular scene text. Pytorch code can be found at https://github.com/wenwenyu/MASTER-pytorch, and Tensorflow code can be found at https://github.com/jiangxiluning/MASTER-TF.

\end{abstract}



\end{frontmatter}


\section{Introduction}
\begin{figure}[t]
\centering
\includegraphics[width=0.6\columnwidth]{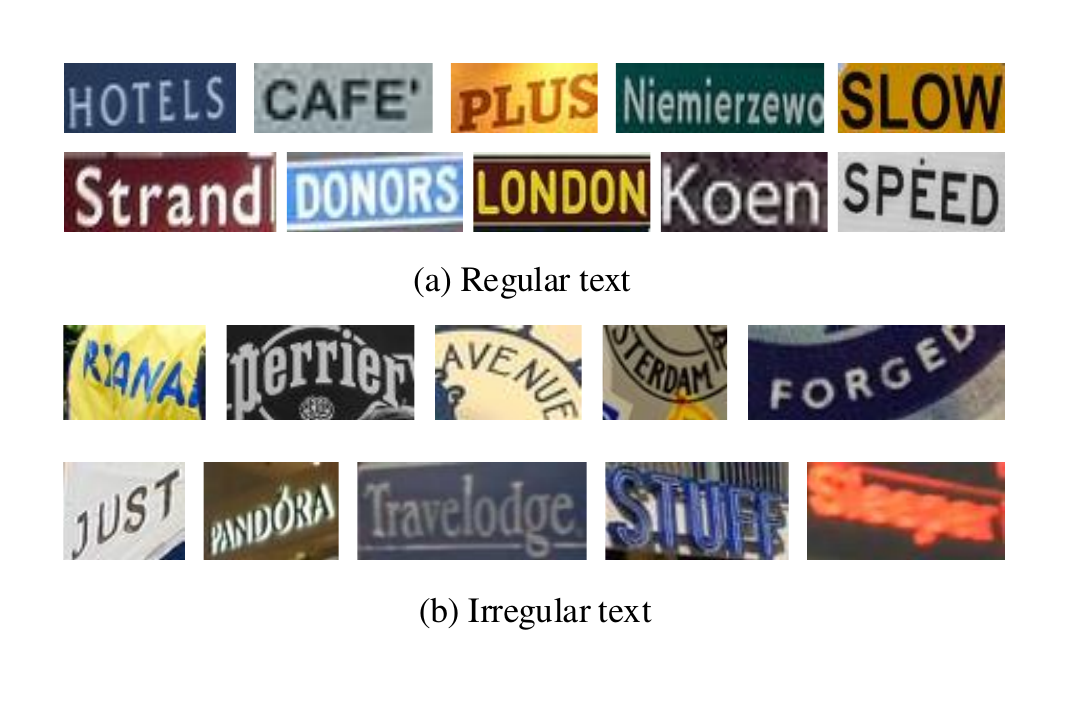} 
\caption{Examples of regular and irregular images. (a). regular text. (b). irregular text.}
\label{fig_regular_and_irregualr_example}
\end{figure}

Scene text recognition in the wild is a hot area in both industry and academia in the last two decades~\cite{Gu2015RecentAI, Liu2018SpecialIO, Van2019APB}. There are various application scenarios such as text identification on the signboard for autonomous driving, ID card scan for a bank, and key information extraction in Robotic Process Automation (RPA). However,  constructing a high-quality scene text recognition system is a non-trivial task due to unexpected blur, strong exposure, spatial and perspective distortion,  and complex background. There are two types of scene text in nature, \textbf{regular} and \textbf{irregular}, as exemplified in Figure \ref{fig_regular_and_irregualr_example}.

Regular scene text recognition aims to recognize a sequence of characters from an almost straight text image. It is usually considered as an image-based sequence recognition problem. Some traditional text recognition methods~\cite{Shivakumara2011} use human-designed features to segment patches into small glyphs, then categorize them into corresponding characters. However, these methods are known to be vulnerable to the complicated background, diverse font types, and irregular arrangement of the characters. Connectionist temporal classification (CTC) based methods~\cite{Shi2015AnET,Bai2018EditPF} and attention-based methods~\cite{Shi2018ASTERAA,Li2018ShowAA}  are the mainstream methods for scene text recognition because they do not require character-level annotations and also show superior performance on real applications.
  
Irregular scene text recognition is more challenging due to various curved shapes and perspective distortions. Existing irregular scene text recognizers can be divided into three categories: rectification based, multi-direction encoding based, and attention-based approaches. Shi \emph{et al.}~\cite{Shi2018ASTERAA} propose ASTER to combine a Thin-Plate Spline (TPS)~\cite{TPS24792} transformation as rectification module and an attentional BiLSTMs as recognition module. ASTER achieves excellent performance on many public benchmarks. Yang \emph{et al.}~\cite{Yang2019SymmetryConstrainedRN} put forward a Symmetry-constrained Rectification Network (ScRN) to tackle highly curved or distorted text instances. Liao \emph{et al.}~\cite{Liao2019SceneTR} conduct scene text recognition from a two-dimensional perspective. Xie \emph{et al.}~\cite{Xie2019ConvolutionalAN} utilize a convolutional attention networks for unconstrained scene text recognition. Fang \emph{et al.}~\cite{Fang2018AttentionAL} propose to ensemble attention and language models in an attention-based architecture. Inspired by the Show-Attend-Tell \cite{Xu2014}, Li \emph{et al.}~\cite{Li2018ShowAA} propose a Show-Attend-Read (SAR) method which employs a 2D attention in encoder-decoder architecture. Nonetheless, attention drifting remains a serious problem in these methods, especially when the text lines contain repetitive digits or characters.
  
Incorporating global context proves to be an effective way to alleviate the problem of attention drifting. Self-attention \cite{Vaswani2017AttentionIA} provides an effective approach to encode global context information. Recently, self-attention attracts a lot of eyeballs
and gains unprecedented success in natural language processing~\cite{Vaswani2017AttentionIA,Devlin2018BERT,yang2019xlnet,gehring2017convolutional} and computer vision~\cite{Cao2019GCNetNN,Hu2017RelationDetection}. It can depict a long-range dependency between words and image regions. Wang \emph{et al.}~\cite{Wang_2018} proposes a Transformer-like non-local block that can be plugged
in any backbone to model spatial global dependencies between objects. Its successors, GCNet, proposed in~\cite{Cao2019GCNetNN}, found that the attention maps are almost the same for different query positions. GCNet simplifies non-local block with SE block \cite{Hu2018} to reduce the computational complexity and enhances the representative ability of the proposed block based on a query-independent formulation.
   
Inspired by the effectiveness of the global context in GCNet and the huge success of the Transformer achieved in NLP and CV, we propose a \textbf{M}ulti-\textbf{A}spect non-local network for irregular \textbf{S}cene \textbf{TE}xt \textbf{R}ecognition (MASTER) to target an efficient and accurate scene text recognition for both regular and irregular text. Our main contributions are highlighted as follows:
\begin{itemize}
\item We propose a novel multi-aspect non-local block and fuse it into the conventional CNN backbone, which enables the feature extracter to model a global context. The proposed multi-aspect non-local block can learn different aspects of spatial 2D attention, which can be viewed as a multi-head self-attention module. Different types of attention focus on different aspects of spatial feature dependencies, which is another form of different syntactic dependency types.
\item In the inference stage, we introduce a memory-cache based decoding strategy to speed up the decoding procedure. The primary means are to remove unnecessary computation and cache some intermediate results of previous decoding times.
\item Besides of its high efficiency, our method achieves the state of the art performance on both regular and irregular scene text benchmarks. Especially, our method achieves the best case-sensitive performance on the COCO-Text dataset.
\end{itemize}

\section{Related Works}

In academia, scene text recognition can be divided into two categories: regular and irregular texts. In this section, we will give a brief review of related works in both areas. A more detailed review for scene text detection and recognition can be found in~\cite{6945320,Zhu2016, Zhang2017OnlineAO, Giotis2017ASO}. 

\emph{Regular text recognition} attracts most of the early research attention. Mishra \emph{et al.}~\cite{Mishra_2016} use a traditional sliding window-based method to describe bottom-up cues and use vocabulary prior to model top-down cues. These cues are combined to minimize the character combination's energy. Shi \emph{et al.}~\cite{Shi2015AnET}  propose an end-to-end trainable character annotation-free network, called CRNN. CRNN extracts a 1D feature sequence using CNN and then encodes the sequence encoding using RNN. Finally, a CTC loss is calculated. CTC loss only needs word-level annotation instead of character-level annotation. Su \emph{et al.}\cite{Su2017AccurateRO} also proposed a method performing word-level recognition without character segmentation using a recurrent neural network. Bigorda \emph{et al.}~\cite{Bigorda2016TextProposalsAT} design a text-specific selective search algorithm to generates a hierarchy of word hypotheses for word spotting in the wild. Gao \emph{et al.}~\cite{Gao2019} integrates an attention module into the residual block to amplify the response of the foreground and suppress the response of the background. However, the attention module cannot encode global dependencies between pixels. Cheng \emph{et al.}~\cite{Cheng2017FocusingAT} observe that attention may drift due to the complex scenes or low-quality images, which is a weakness of the vanilla 2D-attention network. To address the misalignment between the input sequence and the target, Bai \emph{et al.}~\cite{Bai2018EditPF} employs an attention-based encoder-decoder architecture, and estimate the edit probability of a text conditioned on the output sequence. Edit probability is to target the issue of character missing and superfluous. Zhang \emph{et al.}~\cite{Zhang2019a} adopts an unsupervised fixed-length domain adaptation methodology to a variable-length scene text recognition area and the model is also based on attentional encoder-decoder architecture.

\emph{Irregular text recognition} is more challenging than regular text recognition, nevertheless, it appeals to most of researchers' 
endeavour. Shi \emph{et al.}~\cite{ShiWLYB16,Shi2018ASTERAA}  attempt to address the multi-type irregular text recognition problem in one framework via Spatial Transformer Network (STN)~\cite{Jaderberg2015SpatialTN}. In \cite{Zhan_2019_CVPR}, Zhan \emph{et al.} propose to iteratively rectify text images to be fronto-parallel to further improve the recognition performance. The proposed line-fitting transformation estimates the pose of the text line by learning a middle line of the text line and $L$ line segments that are required by Thin-Plate Spline. However, the rectification-based methods are often constrained by the characters' geometric features and the background noise could be exaggerated unexpectedly. To overcome this, Luo \emph{et al.}~\cite{Luo2019MORANAM} propose a multi-object rectified attention network which is more flexible than direct affine transformation estimation. Unlike the rectification-based approaches,  Show-Attend-Read (SAR) proposed by Li~\cite{Li2018ShowAA} uses a 2D-attention mechanism to guide the encoder-decoder recognition module to focus on the corresponding character region. This method is free to complex spatial transformation.

While 2D attention can represent the relationship between target output and input image feature, the global context between pixels and the latent dependency between characters is ignored. In 
\cite{Hu2017RelationDetection}, Hu \emph{et al.} proposes an object relation module to simultaneously model a set of object relations through their visual features. After the success of Transformer~\cite{Vaswani2017AttentionIA},  Wang \emph{et al.} \cite{Wang_2018} incorporate a self-attention block into non-local network. Cao \emph{et al.}~\cite{Cao2019GCNetNN} further simplify and improve the non-local network, and propose a novel global context network (GCNet). Recently, Sheng \emph{et al.}~\cite{Sheng2018NRTRAN} propose a purely Transformer-based scene text recognizer that can learn the self-attention of encoder and decoder. It extracts a 1D sequence feature using a simple CNN module and inputs it into a Transformer to decode target outputs. Nevertheless, the self-attention module of the Transformer consists of multiple fully connected layers, which largely increases the number of parameters. Lee \emph{et al.}~\cite{Lee2020OnRT}, use the self-attention mechanism to 
capture two-dimensional (2D) spatial dependencies of characters. A locality-aware
feedforward layer is introduced in their encoder. Wang \emph{et al.}~\cite{Yang2019AHR} directly abandon the encoder of the original Transformer and only retain the CNN feature extractor and decoder to conduct an irregular scene text recognizer. However, it cannot encode the global context of pixels in the feature map. The network proposed in this paper learns not only the 2D attention between the input feature and output target but also the self-attention inside the feature extractor and decoder. The multi-aspect non-local block can encode different types of spatial feature dependencies with lower computational cost and a compact model.

\section{Methodology}
\begin{figure*}[htbp]
\centering
\includegraphics[width=0.99\textwidth]{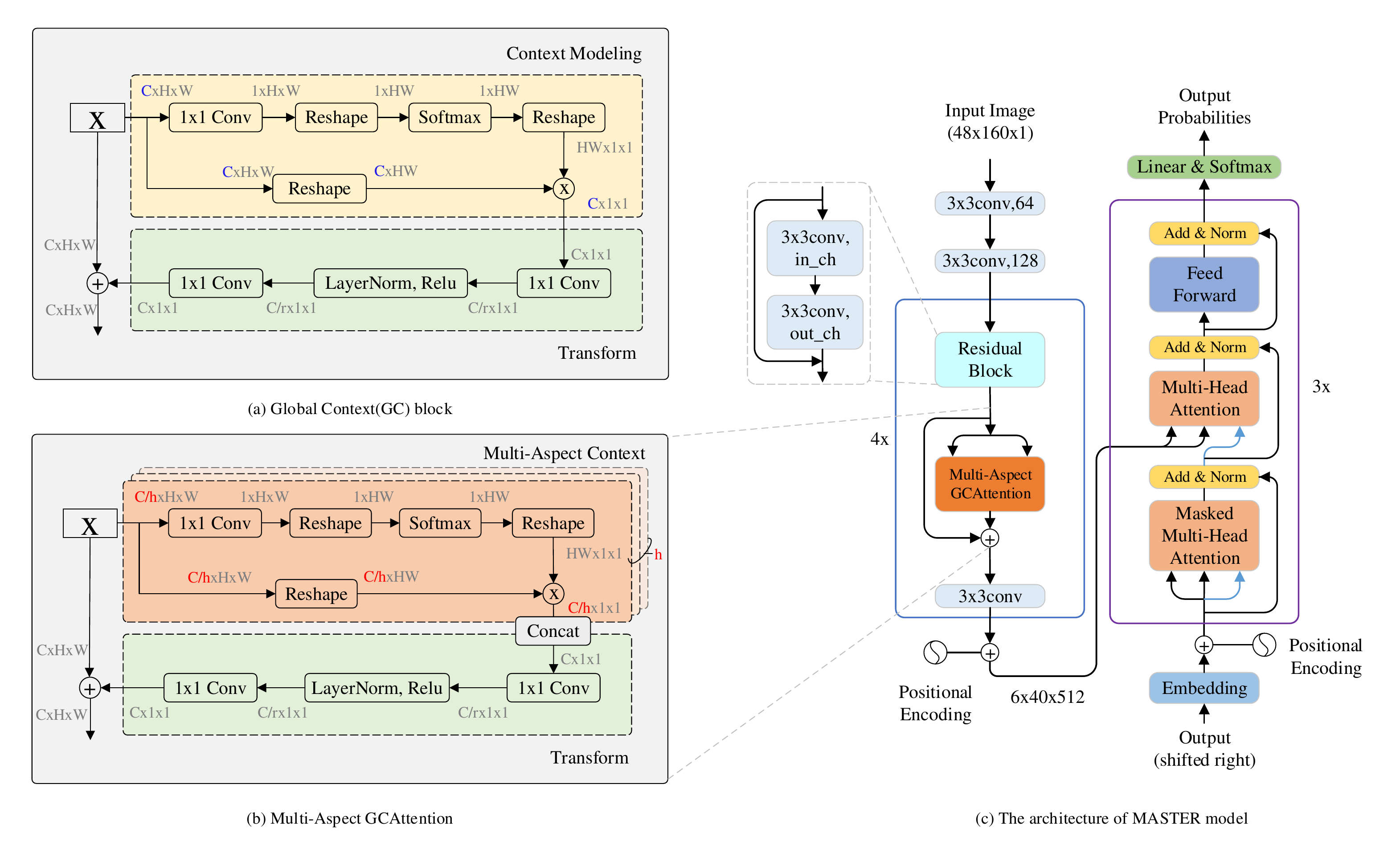}
\caption{(a) representing the architecture of a standard Global Context(GC) block. (b) representing the proposed Multi-Aspect GCAttention.  (c) representing the whole architecture of MASTER model, consisting of two
main parts: a Multi-Aspect Global Context Attention(GCAttention) based encoder for feature representation and a transformer based decoder model. C$\times{H \times{W}}$ denotes a feature map with channel number C, height H and width W. $h$, $r$, and $C/r$  denotes the number of Multi-Aspect Context, bottleneck ratio and hidden representation dimension of the bottleneck, respectively. $\otimes$ denotes matrix multiplication, $\oplus$ denotes broadcast element-wise addition. in\_ch/out\_ch donates input/output dimensions. }
\label{core_unit_and_architerture}
\end{figure*}

MASTER model, as shown in Figure~\ref{core_unit_and_architerture}c, consists of two key modules, a Multi-Aspect Global Context Attention (GCAttention) based encoder and a Transformer based decoder. In MASTER, an image with fixed size is input into the network, and the output is a sequence of predicted characters.
\subsection{Encoder}
Encoder, in our MASTER model, encodes an input image into a feature tensor. For instance, we can obtain a $ 6 \times{40\times512} $ tensor when inputting a $48\times{160\times1}$ image into the encoder of MASTER. One of our key contribution in this paper is that we introduce a Multi-Aspect Global Context Attention (GCAttention) in the encoder part. In this subsection, we will review the definition of the Global Context Block~\cite{Cao2019GCNetNN}, and then introduce the proposed Multi-Aspect Global Context Attention (GCAttention), and finally describe the architecture of the encoder in detail.

\subsubsection{Global Context Block} A standard global context block was firstly introduced in~\cite{Cao2019GCNetNN}. The module structure is shown as Figure~\ref{core_unit_and_architerture}a.  From Figure~\ref{core_unit_and_architerture}a, the input feature map of global context block is ${\bf x}=\{{\bf x}_i\}_{i=1}^{H \times W} \in \mathcal{R}^{C \times H \times W}$ ($C= d_{model}$), where $C$, $W$, and $H$ indicate the number of channel, width and height of the feature map individually. $d_{model}$ indicates the dimension of the output of the encoder. In global context block, three operations are performed on the feature map ${\bf x}$, including (a) global attention pooling for \textbf{context modeling}, (b) bottleneck \textbf{transform} to capture channel-wise dependencies, and (c) broadcasting element-wise \textbf{addition} for feature fusion. The global context block can be expressed as
\begin{equation}
{{\bf{y}}_i} = {{\bf{x}}_i} + {{\bf w}_{v2}}{\rm{ReLU}}\bigg( {{\rm{LN}}\bigg( {{{\bf w}_{v1}}\sum\limits_{\forall j} {\frac{{{e^{{{\bf w}_k}{{\bf{x}}_j}}}}}{{\sum\limits_{\forall m} {{e^{{{\bf w}_k}{{\bf{x}}_m}}}} }}{{\bf{x}}_j}} } \bigg)} \bigg)\,,
\end{equation}
where ${\bf x}$ and ${\bf y}$ denote the input and the output of the global context block, respectively. They have the same dimensions. $i$ is the index of query positions, $j$ and $m$ enumerates positions of all pixels. ${\bf w}_{v1}$, ${\bf w}_{v2}$ and ${\bf w}_{k}$ denote linear transformations to be learned via a $1 \times 1$ convolution. $\rm{LN}(\cdot)$ denotes layer normalization as \cite{Ba2016LayerN}. For simplification, we denote $\alpha_j=\frac{{{e^{{{\bf w}_k}{{\bf{x}}_j}}}}}{{\sum\nolimits_m {{e^{{{\bf w}_k}{{\bf{x}}_m}}}} }}$ as the weight for \textbf{context modeling}, and $\delta(\cdot)={\bf w}_{v2}{\rm{ReLU}}({\rm{LN}}({\bf w}_{v1}(\cdot)))$ as the bottleneck \textbf{transform}. ``$+$'' operation denotes broadcast element-wise \textbf{addition}.

In the Global Context Block, as shown in Figure~\ref{core_unit_and_architerture}a, the softmax operation follows behind a 1x1 Conv and flatten operation, in which the feature map will be converted from (C, H, W) to (1, H*W). The generated vector (1, H*W) is a channel-agnostic feature and it captures spatial information of feature map. Besides, the softmax operation depicts a long-range dependency between pixels in the feature map.

\subsubsection{Multi-Aspect GCAttention} Instead of performing a single attention function in original global context block, we found it beneficial to multiple attention function. Here, we call it as Multi-Apsect GCAttention (MAGC). The structure of the MAGC is illustrated in Figure~\ref{core_unit_and_architerture}b, and we can formulate MAGC as
\begin{equation}
\left\{ \begin{array}{l}
\displaystyle {{\bf{y}}} = {{\bf{x}}} + \delta (MAGC({\bf{x}}) )\,,	\\
\displaystyle MAGC({\bf{x}}) = Concat(gc_1, gc_2, \dots, gc_h)\,,	\\
\displaystyle gc_i =  \sum\limits_{j=1}^{L}	\alpha_j {\bf{x}}_j \,, \\
\displaystyle {\bm{\alpha} }=softmax\left( \frac{{{{\bf w}_k}{{\bf{x}}_1}}}{\sqrt{d_h}},\frac{{{{\bf w}_k}{{\bf{x}}_2}}}{\sqrt{d_h}}, \ldots, \frac{{{{\bf w}_k}{{\bf{x}}_{L}}}}{\sqrt{d_h}}\right)\,, \\
\end{array} \right. 
\end{equation}
where $ h $ is the number of Multi-Aspect Context, $gc_{i}$ denotes the $i$-th global context, $L$ is the number of positions of all pixels in the feature map ($L=W \times H$), $Concat(\cdot)$ is a concatenation function. $MAGC(\cdot)$  denotes multi-aspect global context attention operation. $ \sqrt{d_h} $ is a scale factor to counteract the effect of different variance in MAGC. It can be calculated as  $ d_h = \frac{d_{model}}{h}$.

\subsubsection{Encoder Structure} The detailed architecture of Multi-Aspect GCAttention based Encoder is shown in the left half of Figure \ref{core_unit_and_architerture}c. The backbone of the encoder, following the design of ResNet31 \cite{He2015DeepRL} and the setting protocol in \cite{Li2018ShowAA}, is presented in Table \ref{table_backbone}. The encoder has four fundamental blocks shown in blue color in Figure~\ref{core_unit_and_architerture}c, each fundamental block consists of a residual block, a MAGC, and a convolution block, and max pooling that is not included in the last two fundamental blocks. In the residual block, if the input and output dimensions are different we use the projection shortcut, otherwise, we use the identity shortcut. After the residual block, a Multi-Aspect GCAttention is plugged into network architectures to learn new feature representation from multi-aspect. All the convolutional kernel size is  $ 3\times3 $ . Besides two $ 2\times2 $ max-pooling layers, we also use a $ 1\times2 $ max-pooling layer, which reserves more information along the horizontal axis and benefits the recognition of narrow shaped characters.

\subsection{Decoder}
As shown in the right halves of Figure \ref{core_unit_and_architerture}c, the decoder contains a stack of  $N = 3$ fundamental blocks as shown in purple color. Each fundamental block contains three core modules, a Masked Multi-Head Attention, a Multi-Head Attention, and a Feed-Forward Network (FFN). In the following, we introduce these three key modules in detail, then discuss the loss function used in this paper, and finally introduce memory-cache based inference mechanism.

\subsubsection{Scaled Multi-Head Dot-Product  Attention}
A scaled multi-head dot product attention is firstly introduced in [10]. The inputs of the scaled dot-product attention consist of a query  $ \mathbf{q}_i^T  \in \mathcal{R}^{d} , i \in {[1, t^\prime]}$, (where $ d = d_{model} $ is the dimension of embedding output and $t^\prime $ is the number of queries), and a set of key-value pairs of $d$-dimensional vectors $\left\{\left(\mathbf{k}_{j}, \mathbf{v}_{j}\right)\right\}_{j \in { [1, t] }}$ , $ \mathbf{k}_{j}^T \in {\mathcal{R}^d}$, $ \mathbf{v}_{j}^T \in {\mathcal{R}^d}$ (where $t$ is the number of key-value pairs). 
The formulation of scaled dot-product attention can be expressed as follows
\small
\begin{equation}
\left\{ \begin{array}{l}
\displaystyle Atten(\mathbf{Q}, \mathbf{K}, \mathbf{V})=\left[\mathbf{a}_{1}, \mathbf{a}_{2}, \ldots, \mathbf{a}_{t^{\prime}}\right]^T \in \mathcal{R}^{ t^{\prime} \times d}\,, \\

\displaystyle \mathbf{a}_{i}=Atten\left(\mathbf{q}_{\mathbf{i}}, \mathbf{K}, \mathbf{V}\right)\,,\\

\displaystyle Atten(\mathbf{q_i}, \mathbf{K}, \mathbf{V})=\sum_{j=1}^{t} \alpha_{j} \mathbf{v}_{j}^T \in \mathcal{R}^{d}\,,	\\

\displaystyle \bm{\alpha}= softmax\left(\frac{\left\langle\mathbf{q_i}, \mathbf{k}_{1}^T\right\rangle}{\sqrt{d}}, \frac{\left\langle\mathbf{q_i}, \mathbf{k}_{2}^T\right\rangle}{\sqrt{d}}, \ldots, \frac{\left\langle\mathbf{q_i}, \mathbf{k}_{t}^T\right\rangle}{\sqrt{d}}\right)\,, \\
 
\end{array} \right. 
\end{equation}
\small
where $ \bm{\alpha}$ is the attention weights, and $\mathbf{K}=\left[\mathbf{k}_{1}; \mathbf{k}_{2}; \ldots; \mathbf{k}_{t}\right] \in {\mathcal{R}^{t \times d}}$, $\mathbf{V}= \left[\mathbf{v}_{1}; \mathbf{v}_{2}; \ldots; \mathbf{v}_{t}\right] \in {\mathcal{R}^{t \times d}}$. $\mathbf{Q}=\left[\mathbf{q}_{1}; \mathbf{q}_{2}; \ldots; \mathbf{q}_{t^{\prime}} \right] \in {\mathcal{R}^{t^{\prime} \times d}}$ is a set of queries.

The above scaled dot-product attention can be repeated multiple times (multi-head) with different linear transformations on $\mathbf{Q}$, $\mathbf{K}$ and $\mathbf{V}$, followed by a concatenation and linear transformation operation:
\begin{small}
\begin{equation}
MHA(\mathbf{Q}, \mathbf{K}, \mathbf{V})=\left[\mathbf{head}_{1} , \ldots , \mathbf{head}_{H}\right]\mathbf{W}^{o} \in \mathcal{R}^{t^{\prime}  \times d }\,,
\label{eq_dotprod3}
\end{equation}
\end{small}where $\mathbf{head}_{i}=Atten\left( \mathbf{Q} \mathbf{W}_{i}^{q}, \mathbf{K} \mathbf{W}_{i}^{k}, \mathbf{V}\mathbf{W}_{i}^{v} \right) \in {\mathcal{R}^{t^\prime \times \frac{d}{H}}}$, $MHA(\cdot)$  denotes multi-head attention operation. The parameters are $\mathbf{W}_{i}^{q} \in \mathcal{R}^{d \times\frac{d}{H} }, \mathbf{W}_{i}^{k} \in \mathcal{R}^{d \times  \frac{d}{H}}, \mathbf{W}_{i}^{v} \in$
$\mathcal{R}^{ d\times \frac{d}{H}}$ and $\mathbf{W}^{o} \in \mathcal{R}^{d \times d} $. $H$ denotes the number of multi-head attention.


\subsubsection{Masked Multi-Head Attention} 
This module is identical to the decoder of Transformer~\cite{Vaswani2017AttentionIA}. Masked multi-head attention is an effective mechanism to promise that, in the decoder, the prediction of one time step $t$ can only access the output information of its previous time steps. In the training stage, by creating a lower triangle mask matrix, the decoder can output predictions for all time steps simultaneously instead of one by one sequentially. This makes the training process highly parallel.

\subsubsection{Position-wise Feed-Forward Network} Point-wise Feed-Forward Network (FFN) consists of two fully connected layers. Between these two layers, there is one ReLU activation function. FFN is defined as
\begin{equation}
FFN(\bf x)=\max\left(0, {\bf x} \mathbf{W}_{1} + b_{1}\right)\mathbf{W}_{2} + b_{2},
\label{eq_ffn}
\end{equation}
where the weights are $\mathbf{W}_{1} \in  \mathcal{R}^{d \times d_{f f}} $ , and $\mathbf{W}_{2} \in \mathcal{R}^{ d_{ff} \times d}$ , and the bias are $\mathbf{b}_{1} \in \mathcal{R}^{ d_{ff} }$ and $\mathbf{b}_{2} \in \mathcal{R}^{ d}$, $d_{ff}$ is the inner-dimension of the two linear transformations. The aim of this module is to bring in more non-linearity to the network.

\begin{algorithm}[htbp]
\small
{
  \SetKwRepeat{Do}{do}{while}%
 \SetKwFunction{MaskedMHA}{{MaskedMHA}}
  \SetKwFunction{MHA}{MHA}
  \SetKwFunction{FeedForward}{FeedForward}
  \SetKwFunction{LinearSoftmax}{LinearSoftmax}
  \SetKwFunction{Argmax}{Argmax}
  \SetKwFunction{Concat}{Concat}
   \SetKwFunction{Embedding}{Embedding}
    \SetKwFunction{PositionEmbedding}{PositionEmbedding}
  \SetKwInOut{Input}{Input}\SetKwInOut{Output}{Output}\SetKwInOut{Cached}{Cached Variables}
\SetKwInOut{Params}{params}\SetKwInOut{Initialization}{Initialization}
  \Input{ CNN feature: \textit{F} }
  \Output{$outputs$}
  
     \For{$b\ in \ range(B)$}
     {
     $X^b_k, X^b_v = W^b_k * F,  W^b_v * F$; 
     
     $keys\_memory[b], values\_memory[b] = [\ ], [\ ];$ 
     
     } 
  
	 $t \leftarrow$ 0\;
	 $outputs = [\ ]$\;
	 $ p_t \leftarrow <$SOS$> $\;
	\While{ $p_t \neq <$EOS$>$ and $t\leq T$}
	{
	    $q$ = \Embedding($p_t$) + \PositionEmbedding($t$);
	    

		\For{$b\ in \ range(B) $}
		{
		$keys\_memory[b].append(M^b_k * q);$ 
		
		$values\_memory[b].append(M^b_v * q);$
		
		$q \leftarrow$ \MaskedMHA{$M^b_q * q, keys\_memory[b], values\_memory[b]$}; 
		
		$q \leftarrow$ \MHA{$W^b_q * q$, $X^b_k, X^b_v$}; 
		
		$q \leftarrow$ \FeedForward{q};
		
		}
		
		$t\leftarrow t+1$\;	
		
		$p_t \leftarrow \Argmax{\LinearSoftmax{q}}$\;
		
		$outputs.append(p_t)$
	
  	}
  \caption{Memory-Cache based Inference. $B$ is the number of blocks. $\textit{F}$ is the addition of the CNN feature and the position embedding feature. $T$ is the max decoder length. $\textit{M}$ and $\textit{W}$ are the parameters of the masked multi-head and multi-head attention. $X^b_k$, $X^b_v$, $keys\_memory$, $values\_memory$ are the cached variables.}
  \label{mcd_pseudo}
  
}

\end{algorithm}

\subsubsection{Loss Function}
A linear transformation followed by a softmax function is used to compute the prediction probability over all classes. Then, we use the standard cross-entropy to calculate the loss between the predicted probabilities w.r.t. the ground truth, at each decoding position. In this paper, we use 66 symbol classes except for COCO-Text which uses 104 symbol classes. These 66 symbols are 10 digits, 52 case-sensitive letters, and 4 special punctuation characters. These 4 special punctuation characters are ``$<$SOS$>$'', ``$<$EOS$>$'', ``$<$PAD$>$'', and ``$<$UNK$>$'' which indicate the start of the sequence, the end of the sequence, padding symbol and unknown characters (that are neither digit nor character), respectively. The parameters of the classification layer are shared over all decoding positions.

\subsection{Memory-Cache based Inference Mechanism}
The inference stage is different from the training stage. In the training stage, by constructing a lower triangular mask matrix,  the decoder can predict out all-time steps simultaneously. This process is highly parallel and efficient, where parallel means the batch mechanism. However, the decoder in the inference stage can only predict each character one by one sequentially until the decoder predicts out the ``EOS'' token or the length of the decoder sequence reaches to the maximum length. In the inference stage, the output of the later step is dependent on the outputs of its previous time steps because the outputs of its previous time steps will be used as part of the input to decode itself.

To speed up the decoding process, we introduce a new decoding mechanism named memory-cache based decoding inspired by XLNet~\cite{yang2019xlnet}. The memory-cache based decoding strategy is described in Algorithm~\ref{mcd_pseudo} in pseudo-code. The primary approaches are to cache some intermediate results of previous decoding times in Lines 2 and 11-12, and to remove unnecessary computation in Lines 13-14 of Algorithm~\ref{mcd_pseudo}.  
In each decoding step, $q$ is always a 1D vector instead of a 2D matrix in traditional decoding framework.



\section{Experiments}
We conduct extensive experiments on several benchmarks to verify the effectiveness of our method and compare it with the state-of-the-art methods. In Section~\ref{datasets}, we give an introduction to the used training and testing datasets. Then in Section~\ref{Implementationdeatils}, we present our implementation details. In Section~\ref{comparisons}, we make a detailed comparison between our method and the state-of-the-art methods. Finally, we conduct an ablation study in Section~\ref{Ablation}.
\begin{table}[htbp]
\centering
\caption{{A ResNet-based CNN network architecture for robust text feature representation. Residual blocks are shown in brackets, and Multi-Aspect GCAttention is highlighted with gray background.  ``$3 \times 3, 1 \times 1, 1 \times 1, 128$'' denotes the kernel size, the stride, the padding, and the output channel of a convolution layer respectively}. The ``Output'' column means the spatial shape $height \times width$ of the output.
}
 
\resizebox{.45\columnwidth}{!}{
\begin{tabular}{| c | c | c |}
\hline
\textbf{Layer}   & \textbf{Configuration}  &\textbf{ Output}  \\
\hline
\multirow{3}{*}{conv1\_x}& $3\times3$, $1\times1$, $1\times1$, 64  & $48\times160$ \\ \cline{2-3}
                         & $3\times3$, $1\times1$, $1\times1$, 128 & $48\times160$ \\\cline{2-3}
                         & max\_pool: $2\times2$, $2\times2$, $0\times0$ & $24\times80$ \\
\hline
\multirow{4}{*}{conv2\_x}&  $\left[ \begin{array}{c}
							 3\times3, 256\\
							 3\times3, 256\\
						 \end{array} \right]\times 1 $
                         & $24\times80$ \\ \cline{2-3}
                         & \cellcolor{black!5} multi-aspect gcattention & $24\times80$\\\cline{2-3}
                         & $3\times3$, $1\times1$, $1\times1$, 256 & $24\times80$ \\\cline{2-3}
                         & max\_pool: $2\times2$, $2\times2$, $0\times0$ & $12\times40$  \\
\hline  
                     
\multirow{4}{*}{conv3\_x}&  $\left[ \begin{array}{c}
							 3\times3, 512\\
							 3\times3, 512\\
						 \end{array} \right]\times 2 $
                         & $12\times40$ \\ \cline{2-3}
                         & \cellcolor{black!5} multi-aspect gcattention & $12\times40$\\\cline{2-3}
                         & $3\times3$, $1\times1$, $1\times1$, 512 & $12\times40$ \\\cline{2-3}
                         & max\_pool: $2\times1$, $2\times1$, $0\times0$ & $6\times40$  \\    
\hline
                     
\multirow{4}{*}{conv4\_x}&  $\left[ \begin{array}{c}
							 3\times3, 512\\
							 3\times3, 512\\
						 \end{array} \right]\times 5 $
                         & $6\times40$ \\ \cline{2-3}
                         &  \cellcolor{black!5}multi-aspect gcattention & $6\times40$\\\cline{2-3}
                         & $3\times3$, $1\times1$, $1\times1$, 512 & $6\times40$ \\ 
\hline
                     
\multirow{4}{*}{conv5\_x}&  $\left[ \begin{array}{c}
							 3\times3, 512\\
							 3\times3, 512\\
						 \end{array} \right]\times 3 $
                         & $6\times40$ \\ \cline{2-3}
                         &  \cellcolor{black!5}multi-aspect gcattention & $6\times40$\\\cline{2-3}
                         & $3\times3$, $1\times1$, $1\times1$, 512 & $6\times40$ \\                                          
\hline
\end{tabular}
}
\label{table_backbone}
\end{table}

\subsection{Datasets}
\label{datasets}
In this paper, we train our MASTER model only on three synthetic datasets without any finetuning on other real datasets. We evaluate our model on eight standard benchmarks that contain four regular scene text datasets and four irregular scene text datasets.

The training datasets consist of the following datasets.

\textbf{Synth90k} (MJSynth) is the synthetic text dataset proposed in \cite{Jaderberg14c}. The dataset has 9 million images generated from a set of 90k common English words. Every image in Synth90k is annotated with a word-level ground-truth. All of the images in this dataset are used for training.

\textbf{SynthText}~\cite{Gupta16} is a synthetic text dataset originally introduced for text detection. The generating procedure is similar to~\cite{Jaderberg14c}, but different from~\cite{Jaderberg14c}, words are rendered onto a full image with a large resolution instead of a text line. 800 thousand full images are used as background images, and usually, each rendered image contains around 10 text lines.  Recently, It is also widely used for scene text recognition. We obtain 7 millions of text lines from this dataset for training.

\textbf{SynthAdd} is the synthetic text dataset proposed in \cite{Li2018ShowAA}. The dataset contains 1.6 million word images using the synthetic engine proposed by \cite{Jaderberg14c} to compensate for the lack of special characters like punctuations. All of the images in this dataset are used for training.

The test datasets consist of the following datasets.

\textbf{IIIT5K-Words} (IIIT5K) \cite{Mishra2012TopdownAB} has 3,000 test images collected from the web. Each image contains a short, 50-word lexicon and a long, 1,000-word lexicon. A lexicon includes the ground truth word and other stochastic words.

\textbf{Street View Text} (SVT) \cite{Wang2011EndtoendST} is collected from the Google Street View. The test set includes 647 images of cropped words. Many images in SVT are severely corrupted by noise and blur or have low resolution. Each image contains a 50-word lexicon.

\textbf{ICDAR 2003} (IC03) \cite{Lucas2004ICDAR2R} contains 866 images of the cropped word because we discard images that contain non-alphanumeric characters or have less than three characters for a fair comparison. Each image contains a 50-word lexicon defined.

\textbf{ICDAR 2013} (IC13) \cite{Karatzas2013ICDAR2R} contains 1,095 images for evaluation and 848 cropped image patches for training. We filter words that contain non-alphanumeric characters for a fair comparison, which results in 1,015 test words. No lexicon is provided.

\textbf{ICDAR 2015} (IC15) has 4,468 cropped words for training and 2,077 cropped words for evaluation, which are capture by Google Glasses without careful positioning and focusing. The dataset contains many of irregular text. 

\textbf{SVT-Perspective} (SVTP) consists of 645 cropped images for testing \cite{Phan2013RecognizingTW}. Images are generated from side-view angle snapshots in Google Street View. Therefore, most images are perspective distorted. Each image contains a 50-word lexicon and a full lexicon.

\textbf{CUTE80} (CUTE) contains 288 images \cite{Risnumawan2014ARA}. It is a challenging dataset since there are plenty of images with curved text. No lexicon is provided.

\textbf{COCO-Text} (COCO-T) was firstly introduced in the Robust Reading Challenge of ICDAR 2017. It contains 62,351 image patches cropped from the famous Microsoft COCO dataset. The COCO-T dataset is extremely challenging because the text lines are mixed up with printed, scanned, and handwritten texts, and the shapes of text lines vary a lot. For this dataset, 42,618, 9,896, and 9,837 images are used for training, validation, and testing individually.

\subsection{ Network Structure and Implementation Details}
\label{Implementationdeatils}
\subsubsection{Networks}
The network structure of the Encoder part is listed in Table \ref{table_backbone}. The input size of our model is $48\times160$. When the ratio between width and height is larger than $\frac{160}{48}$, we directly resize the input image into $48\times160$, otherwise, we resize the height to 48 while keeping the aspect ratio and then pad the resized image into to $48\times160$. In MASTER, the embedded dimension  $d$ is 512,  the dimension of the output of the encoder $d_{model}$ is 512 too, and the number $H$ of the multi-head attention is 8. $d_{ff}$ in the feed-forward module is set to be 2048, and the identical layers $N$ is 3. We use 0.2 dropout on the embedding module, feed-forward module, and the output layer of the linear transformation in the decoder part. The number $h$ of Multi-Aspect Context is 8 and the bottleneck ratio  $r$ is 16.

\begin{table*}[htbp]
\centering
\caption{Performance of our model and other state-of-the-art methods on public datasets. All values are reported as a percentage (\%). ``None'' means no lexicon. * indicates using both word-level and character-level annotations to train the model. ** denotes the performance of SAR trained only on the synthetic text datasets. In each column, the best performance result is shown in \textbf{bold} font, and the second-best result is shown with an underline. Our model achieves competitive performance on most of the public datasets, and the distance between us and the first place  \cite{8812908} is very small on IIIT5k and SVT datasets.}
\resizebox{.7\columnwidth}{!}{

\begin{tabular}{| c | c | c | c | c | c | c | c |}
\hline
\multirow{2}{*}{ \textbf{Method} } &  \textbf{IIIT5K} & \textbf{SVT} & \textbf{IC03} & \textbf{IC13} & \textbf{IC15} & \textbf{SVTP} & \textbf{CUTE}   \\
\cline{2-8} 
                                & None & None & None & None & None & None & None \\
\hline
Jaderberg \emph{et al.} \cite{Jaderberg2015SpatialTN}  &  -   & 80.7 & 93.1 & 90.8 & -    &  -   & -     \\
Shi \emph{et al.} \cite{ShiWLYB16}     	& 81.9 & 81.9 & 90.1 & 88.6 & -    & 71.8 & 59.2   \\
STAR-Net \cite{Liu2016STARNetAS}        & 83.3 & 83.6 &  -   & 89.1 & -    & 73.5 & -    \\
Wang and Hu \cite{Wang2017GatedRC}         & 80.8 & 81.5 &  -   &  -   & -    & -    & -    \\
CRNN \cite{Shi2015AnET}             & 81.2 & 82.7 & 91.9 & 89.6 & -    & -    & -    \\
Focusing Attention \cite{Cheng2017FocusingAT}*     & 87.4 & 85.9 & 94.2 & 93.3 & 70.6 & -    & -    \\
SqueezedText \cite{Liu2018SqueezedTextAR}*   & 87.0 & -    & -    & 92.9 & -    & -    & -    \\
Char-Net \cite{Liu2018CharNetAC}*        & 92.0 & 85.5 &  -   & 91.1 & 74.2 & 78.9 & -    \\
Edit Probability \cite{Bai2018EditPF}*           & 88.3 & 87.5 & 94.6 & 94.4 & 73.9 & -    & -    \\
ASTER \cite{Shi2018ASTERAA}          & 93.4 & 89.5 & 94.5 & 91.8	& 76.1 & 78.5 & 79.5    \\
NRTR \cite{Sheng2018NRTRAN}         & 86.5 & 88.3 & \underline{95.4} & \underline{94.7}	& -    & -    & -    \\
SAR** \cite{Li2018ShowAA}           & 91.5 & 84.5 &  -	 & 91.0 & 69.2 & 76.4 & 83.3    \\
ESIR \cite{Zhan_2019_CVPR}          & 93.3 & 90.2 &	 -   & 91.3 & 76.9 & 79.6 & 83.3    \\
MORAN \cite{Luo2019MORANAM}          & 91.2 & 88.3 &	 95.0   & 92.4 & 68.8 & 76.1 & 77.4    \\  
Wang \emph{et al.} \cite{Yang2019AHR}             & 93.3 & 88.1 &	 -   & 91.3	& 74.0 & 80.2 & 85.1    \\
Mask TextSpotter \cite{8812908}*               & \textbf{95.3} & \textbf{91.8} &	 95.2   & \textbf{95.3}	&\underline{ 78.2} & \underline{83.6} & \textbf{88.5}   \\
\hline
 MASTER (Ours)&\underline{95.0} &\underline{90.6}&\textbf{96.4}&\textbf{95.3}&\textbf{79.4}&\textbf{84.5}&\underline{87.5}\\
\hline
\end{tabular}
}
\label{table_performance}
\end{table*}

\subsubsection{Implementation Details}Our model is only trained on three synthetic datasets without any finetune on any real data except for COCO-T dataset. These three synthetic datasets are SynText~\cite{Gupta16} with 7 millions of text images, Synth90K~\cite{Jaderberg14c} with 9 millions of text images and SynthAdd~\cite{Li2018ShowAA} with 1.6 millions of text images. 

Our model is implemented in PyTorch. The model is trained on four NVIDIA Tesla V100 GPUs with $16\times4$ GB  memory. We train the model from scratch using Adam~\cite{Kingma2014AdamAM} optimizer and cross-entropy loss with a batch size of $128\times4$. The learning rate is set to be $4\times 10^{-4}$ over the whole training phase. We observe that the learning rate should be associated with the number of GPUs. For one GPU, $1\times 10^{-4}$ is a good choice. Our model is trained for 12 epochs, each epoch takes about 3 hours.  

Only for COCO-Text,  we further finetune the above model with around 9K real images collected from IC13, IC15, and IIIT5K, and the training and validation images of COCO-Text. 
At the test stage, for the image with its height larger than width, we rotate the images 90 degrees clockwise and anti-clockwise. We feed the original image and two rotated images into the model and choose the output result with the maximum output probability. No lexicon is used in this paper. Different from SAR~\cite{Li2018ShowAA}, ASTER~\cite{Shi2018ASTERAA}, and NRTR~\cite{Sheng2018NRTRAN}, we do not use beam search.

\subsection{Comparisons with State-of-the-arts}
\label{comparisons}
In this section, we measure the proposed method on several regular and irregular text benchmarks and analyze the performance with other state-of-the-art methods. We also report results on the online COCO-Text datasets test server\footnote{\url{https://rrc.cvc.uab.es/?ch=5&com=evaluation&task=2}} to show the performance of our model.

\begin{table*}[htbp]
\centering
\caption{Leaderboard of various methods on the online COCO-Text test server. In each column, \textbf{Bold} represent the best performance.}
\resizebox{.8\columnwidth}{!}{
\begin{tabular}{| c | c | c | c | c |}
\hline
\multirow{2}[10]{*}{\centering \textbf{Method} } & \multicolumn{2}{c|}{\textbf{Case Sensitive}} & \multicolumn{2}{ c |}{\textbf{Case Insensitive}} \\
\cline{2-5}
   & \makecell*[c]{\textbf{Total Edit}\\ \textbf{Distance}}  & \makecell*[c]{\textbf{Correctly}\\ \textbf{Recognised}\\ \textbf{Words (\%)}}  &  \makecell*[c]{\textbf{Total Edit}\\ \textbf{Distance}}  &\makecell*[c]{\textbf{Correctly}\\ \textbf{Recognised}\\ \textbf{Words (\%)}} \\
\hline

	SogouMM      &		3,496.3121 &		44.64 &	\textbf{1,037.2197}    &	\textbf{ \textbf{77.97} } \\
 
SenseTime-CKD      &	4,054.8236 &	41.52 &	824.6449    &	 77.22  \\
HIK\_OCR     	   &	3,661.5785 &	41.72 &	 899.1009	&	76.11 \\
Tencent-DPPR Team  &	4,022.1224 &	36.91 &	 1,233.4609	&	70.83 \\
CLOVA-AI \cite{Baek2019WhatIW}	       &	3,594.4842 & 47.35 &	 1,583.7724	&	69.27 \\
SAR   \cite{Li2018ShowAA}             &	4,002.3563 &	41.27 &	 1,528.7396 &	66.85 \\
HKU-VisionLab \cite{Liu2018CharNetAC}     &	3,921.9388 &	40.17 &	 1,903.3725	&	59.29 \\

\hline
MASTER (single model)& 3,527.3165 & 45.96	 &	1,528.7526 &	67.41\\
MASTER (Ours)& \textbf{3,272.0810} & \textbf{49.09}	 &	 1,203.4201 &	71.33\\
\hline
   
\end{tabular}
}
\label{table_leaderboard}
\end{table*}

As shown in Table \ref{table_performance}, our method achieves superior performance on both regular and irregular datasets compared to the state-of-the-art methods. On the regular datasets including IIIT-5K, IC03, and IC13, our approach largely improves SAR~\cite{Li2018ShowAA} which is based on LSTM with 2D attention and ASTER~\cite{Shi2018ASTERAA} which is based on Seq2Seq model with attention after a text rectification module. Specifically, our approach improves SAR by 3.5\% and 6.1\% on IIIT-5K and SVT individually. On the irregular datasets, our method achieves the best performance on  SVTP and IC15 datasets. This fully demonstrates the multi-aspect mechanism used in MASTER is highly effective in irregular scene text. Note that all these results are not with lexicon and beam search. The method in \cite{8812908} uses extra character-level data.

Furthermore, seen from Table \ref{table_leaderboard}, we also use online evaluation tools on COCO-Text datasets to verify our competitive performance. As we can see, our model outperforms the compared method by a large margin in case sensitive metrics, demonstrating the powerful network. Specifically, our model gets correctly recognised word accuracy increases of 1.74\% (from 47.35\% to 49.09\%) under case sensitive conditions. In the case of case-insensitive metrics, our model also gets the fourth place on the leaderboard and the performance is much better than SAR. Note that, the first place method of case-insensitive uses a tailored 2D-attention module and the second and third place method of case-insensitive leaderboard use model ensemble. Our results are based on ensemble of four models obtained in different time steps of the same round of training process. The prediction with the maximum probability in four models is selected as the final prediction.

\begin{figure}[htb]
\center
\resizebox{.5\columnwidth}{!}{
\begin{tabular}{cccc}

Input Images & Ours & By SAR~\cite{Li2018ShowAA} & GT \\
\hline

\raisebox{-0.5\height}{\includegraphics[width=0.24\linewidth,height=0.7cm]{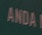}} 
& \makecell*[c]{  \textcolor[rgb]{0,1,0}{ANDA}}
& \makecell*[c]{   \textcolor[rgb]{0,1,0}{A}\textcolor[rgb]{1,0,0}{M}\textcolor[rgb]{0,1,0}{DA}  } & ANDA\\

\raisebox{-0.5\height}{\includegraphics[width=0.24\linewidth,height=0.7cm]{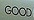}} 
& \makecell*[c]{  \textcolor[rgb]{0,1,0}{GOOD}}
& \makecell*[c]{   \textcolor[rgb]{0,1,0}{G}\textcolor[rgb]{1,0,0}{C}\textcolor[rgb]{0,1,0}{OD}  } & GOOD\\

\raisebox{-0.5\height}{\includegraphics[width=0.24\linewidth,height=0.7cm]{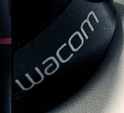}} 
& \makecell*[c]{  \textcolor[rgb]{0,1,0}{wacom}}
& \makecell*[c]{   \textcolor[rgb]{0,1,0}{wac}\textcolor[rgb]{1,0,0}{c}\textcolor[rgb]{0,1,0}{om}  } & wacom\\

\raisebox{-0.5\height}{\includegraphics[width=0.24\linewidth,height=0.7cm]{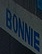}} 
& \makecell*[c]{  \textcolor[rgb]{0,1,0}{BONNIE}}
& \makecell*[c]{   \textcolor[rgb]{0,1,0}{BO}\textcolor[rgb]{1,0,0}{N}\textcolor[rgb]{0,1,0}{IE} } & BONNIE\\

\raisebox{-0.5\height}{\includegraphics[width=0.24\linewidth,height=0.7cm]{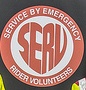}} 
& \makecell*[c]{  \textcolor[rgb]{0,1,0}{SERV}}
& \makecell*[c]{  \textcolor[rgb]{1,0,0}{LEAD} } & SERV\\

\raisebox{-0.5\height}{\includegraphics[width=0.24\linewidth,height=0.7cm]{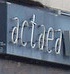}} 
& \makecell*[c]{  \textcolor[rgb]{0,1,0}{actaea}}
& \makecell*[c]{   \textcolor[rgb]{0,1,0}{acta}\textcolor[rgb]{1,0,0}{r}\textcolor[rgb]{0,1,0}{a}  } & actaea\\
\vspace{-1.5em}
\\

\raisebox{-0.5\height}{\includegraphics[width=0.24\linewidth,height=0.7cm]{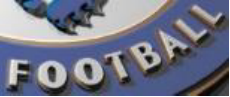}} 
& \makecell*[c]{  \textcolor[rgb]{0,1,0}{FOOTBALL}}
& \makecell*[c]{   \textcolor[rgb]{0,1,0}{FOOTBA}\textcolor[rgb]{1,0,0}{L}\textcolor[rgb]{0,1,0}{}  } & FOOTBALL\\

\raisebox{-0.5\height}{\includegraphics[width=0.24\linewidth,height=0.7cm]{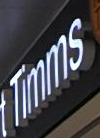}} 
& \makecell*[c]{  \textcolor[rgb]{0,1,0}{Timms}}
& \makecell*[c]{   \textcolor[rgb]{0,1,0}{Ti}\textcolor[rgb]{1,0,0}{mmm}\textcolor[rgb]{0,1,0}{s}  } & Timms\\
\vspace{-1.5em}
\\
\hline
\end{tabular}
}
\caption{Samples of recognition results of our MASTER and SAR. Green characters mean correct predictions and red characters mean wrong predictions. }
\vspace{-.5em}
\label{fig_predict_compare}
\end{figure}

Seen from Figure \ref{fig_predict_compare}, Our method possesses more robust performance on scene text recognition than SAR \cite{Li2018ShowAA}, although the input image quality is blurry and the shape is curved or the text is badly distorted. The reason is that our model not only learns the input-output attention but also learns self-attention which encodes feature-feature and target-target relationships inside the encoder and decoder. This makes the intermediate representations more robust to spatial distortion. Besides, in our approach, the problem of attention drifting is significantly eased. As shown in Figure~\ref{fig_predict_compare}, the attention driftings lead to errors (``FOOTBALL'' and ``TIMMS'' are misrecognized as ``FOOTBAL'' and ``TIMMMS'' individually.) in SAR, but MASTER successfully recognizes these words.

\subsection{Ablation Studies}
\label{Ablation}
\subsubsection{Influence of Key Hyperparameters}
we perform a series of ablation studies to analyze the impact of different hyperparameters on the recognition performance. All models are trained from scratch on three synthetic datasets (Synth90K, SynthText, and SynthAdd). Results are reported on seven standard benchmarks without using a lexicon. Here, we study two key hyperparameters, the number $h$ of Multi-Aspect Context in the encoder part, and the number $N$ of fundamental blocks in the decoder part. The results are shown in Table~\ref{table_ablation}.

\begin{table*}[htb]
\caption{Under different parameter settings our model recognition accuracy: $h$, $N$ denotes the numbers of Multi-Aspect Context in the encoder and identical layers in the decoder, respectively. Standard Setting uses $h=8$ and  $N=3$. When $h$ or $N$ changes, all other parameters keep the same as the Standard Setting. All values are reported as a percentage(\%).}

\centering 
\resizebox{.7\columnwidth}{!}{
\begin{tabular}{|l|c|c|c|c|c|c|c|}
\hline 
\textbf{Methods} & \multicolumn{1}{c|}{\textbf{IIIT5k}} & \multicolumn{1}{c|}{\textbf{SVT}} & \multicolumn{1}{c|}{\textbf{CUTE}}& \multicolumn{1}{c|}{\textbf{IC03}}& \multicolumn{1}{c|}{\textbf{IC13}}& \multicolumn{1}{c|}{\textbf{IC15}}& \multicolumn{1}{c|}{\textbf{SVTP}}\\\hline
\tabincell{l}{Standard Setting: \\ $h=8$, $N=3$} & 95.0  & 90.6 & 87.5& 96.4& 95.3& 79.4& \textbf{84.5}\\
\hline
\hline
$h = 0$ & 94.6 & 90.1 & 86.2& 95.9 & 95.0 & 78.4 & 82.3 \\
$h = 1$ & 94.9 & \textbf{91.5} & 87.6 & \textbf{96.9} & \textbf{95.7} & 79.4 & 83.8 \\
$h = 2$ & 94.93 & 90.7 & \textbf{88.54} & 96.6 & 95.4 & 79.5 & 84.0\\
$h = 4$ & 94.7 & 90.9 & 86.8  & 96.1 & 95.1 & \textbf{79.6} & 83.7\\
$h = 16$ & \textbf{95.1} & 91.3 & 85.4  & 96.0 & 95.3 & 79.4 & 84.1\\
\hline
\hline
$N = 1$ & 94.3 & 90.4 & 85.4 & 95.3 & 94.1 & 78.9 & 83.1 \\
$N = 6$ & 91.3 & 87.4 & 76.7 & 94.3 & 91.6 & 72.9 & 75.7 \\
\hline
\end{tabular}
}
\label{table_ablation}
\end{table*}

There are two groups of experimental comparisons in Table~\ref{table_ablation}. Fix $N=3$, we vary $h$ ranging in $[0, 1,2,4,8,16]$, where $h=0$ means no MAGC is used in the model. We observe that using the MAGC module consistently improves the performance compared to that without using MAGC ($h=0$). Compared to $h=0$, $h=8$ obtains performance improvement on all datasets, especially significant improvement on CUTE, IC15, and SVTP. These three datasets are difficult and irregular. We believe this phenomenon is due to the introduced MAGC module that can well capture different aspects of spatial 2D attention which is very important for irregular and hard text images. We also evaluate different settings $N=[1, 3, 6]$ of the number of fundamental blocks in the decoder part. $N=3$ gets the best performance, and the performance of N=6 decreases a lot compared to $N=3$. We reckon that too deep decoder layers may bring in convergence problems. Therefore, in our experiment, we use $N=3$, $h=8$ in default.

\subsubsection{Comparison of Evaluation Speed}
\begin{table}[t]
\caption{Speed comparison between MASTER (Ours) and SAR. MASTER is faster and more accurate than the SAR method. All timing information is on an NVIDIA Tesla V100 GPU. }

\begin{center}
\small
\resizebox{.7\columnwidth}{!}{
\begin{tabular}{|c|c|c|c|c|}
\hline
  \textbf{Method } &  \textbf{Input}  & \textbf{Accuracy} & \textbf{Inference Time (ms)} &\textbf{Training Time (h)} \\
\hline
SAR \cite{Li2018ShowAA} & $48 \times 160$  & 91.5 & 16.1 & 51 \\
MASTER (original) &  $48 \times 160$  & 95.0  & 9.2 & 36 \\
\hline
MASTER (improved) &  $48 \times 160$  & 95.0  & 4.3 & 36 \\
\hline
\end{tabular}
}
\end{center}
\label{table_speed}
\end{table}

We conduct a comparison of test speed on a server using an NVIDIA Tesla V100 GPU with Intel Xeon Gold 6130@ 2.10 GHz CPU. The results are averaged on 3,000 test images from IIIT-5K, the input image size is $48\times160$. The results of SAR is based on our implementation in PyTorch with the same setting as~\cite{Li2018ShowAA}.

We observe from Table~\ref{table_speed} that, MASTER not only achieves better performance but also runs faster than SAR. By stacking multiple test images together and inputting the stacked batch in one time, we can obtain a speedup. The test time speed of our MASTER is \textbf{9.2 ms per image} compared to 16.1 ms of SAR. By using a new memory-cache based inference mechanism, the decoder can speed up to 4.3 ms from 9.2 ms. Besides, we also compare the training speed between MASTER and SAR. As shown in the last column of Table~\ref{table_speed}. The results show that MASTER has faster training speed because of the parallel training.

\subsubsection{Model stability}

\begin{figure}[htb]
\centering
\includegraphics[width=0.5\columnwidth]{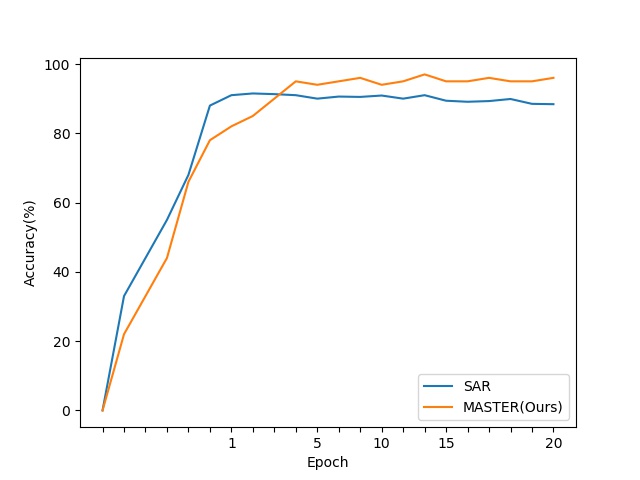} 
\caption{The model stability comparison between MASTER (Ours) and SAR \cite{Li2018ShowAA}. }
\label{fig_model_stability}
\end{figure}

We show the evaluation accuracies of MASTER and SAR along with training steps in Figure~\ref{fig_model_stability}. We find that from Figure~\ref{fig_model_stability}, the MASTER model achieves more stable recognition performance than SAR although SAR converges faster. We reckon the reason is the MASTER requires calculating global attention which is slower but SAR only needs to compute local attention. We can see that the performance of the MASTER model is very stable when it hits the best performance, it will not decrease a lot. However, the performance of SAR often decreases a little more when it reaches the best performance.

\section{Conclusions}
In this work, we propose a novel approach MASTER: multi-aspect non-local network for scene text recognition. The MASTER consists of a Multi-Aspect Global Context Attention (GCAttention) based encoder module and a transformer-based decoder module. The proposed MASTER owns three advantages: (1) The model can both learn input-output attention and self-attention which encodes feature-feature and target-target relationships inside the encoder and the decoder. (2) Experiments demonstrate that the proposed method is more robust to spatial distortions. (3) The training process of the proposed method is highly parallel and efficient, and the inference speed is fast because of the proposed novel memory-cached decoding mechanism. Experiments on standard benchmarks demonstrate it can achieve state-of-the-art performances regarding both efficiency and recognition accuracy.

%

\bibliographystyle{elsarticle-num} 
\bibliography{references,ref}


%
%
%

\end{document}